\documentclass[11pt]{article}
\usepackage{acl2015}
\usepackage{times}
\usepackage{url}
\usepackage{latexsym}
\usepackage{fixltx2e}
\usepackage{graphicx}
\usepackage{hyperref} 
\usepackage{multirow} 
\usepackage{linguex} 
\usepackage{float}
\usepackage{qtree}  

\title{The CLaC Discourse Parser at CoNLL-2015}

\author{Majid Laali \hspace{2cm} Elnaz Davoodi \hspace{2cm} Leila Kosseim \\
  Department of Computer Science and Software Engineering, \\
  Concordia University, Montreal, Quebec, Canada \\
  {\tt \{m\_laali, e\_davoo, kosseim\}@encs.concordia.ca} \\
  }
  
\date{}

\begin{document}
\maketitle
\begin{abstract}
   This paper describes our submission (\textit{kosseim15}) to the CoNLL-2015 shared task on shallow discourse parsing. We used the UIMA framework to develop our parser and used ClearTK to add machine learning functionality to the UIMA framework. 
   Overall, our parser achieves a result of 17.3 F\textsubscript{1} on the identification of discourse relations on the blind CoNLL-2015 test set, ranking in sixth place. 
\end{abstract}

\section{Introduction}
\begin{figure*}[ht]
\centering
\includegraphics[height=13mm]{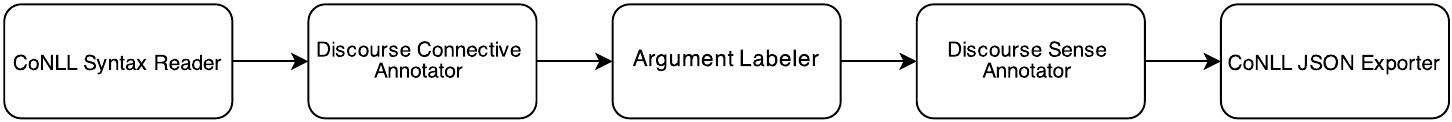}
\caption{Components of the CLaC Discourse Parser}
\label{fig:architecture}
\end{figure*}


Today, discourse parsers typically consist of several independent components that address the following problems:
\begin{enumerate}
\item \textit{Discourse Connective Classification}: The concern of this problem is the identification of discourse usage of discourse connectives within a text. 
\item \textit{Argument Labeling}: This problem focuses on labeling the text spans of the two discourse arguments, namely \textsc{Arg1} and \textsc{Arg2}.
\item \textit{Explicit Sense Classification}: This problem can be reduced to the sense disambiguation of the discourse connective in an explicit discourse relation.
\item \textit{Non-Explicit Sense Classification}: The target of this problem is the identification of implicit discourse relations between two consecutive sentences.
\end{enumerate}

To illustrate these tasks, consider Example~\ref{ex:one}:
\ex. \label{ex:one} \textit{We would stop index arbitrage} \underline{when} \textbf{the market is under stress}.\footnote{The example is taken from the CoNLL 2015 trial dataset.}

The task of \textit{Discourse Connective Classification} is to determine if the marker \textit{``when''} is used to mark a discourse relation or not. \textit{Argument Labeling} should segment the two arguments \textsc{Arg1} and \textsc{Arg2} (in this example, \textit{\textsc{Arg1} is italicized} while \textbf{\textsc{Arg2} is bolded}). Finally, \textit{Explicit Sense Classification} should identify which discourse relation is signaled by \textit{``when''} - in this case \textsc{Contingency.Condition}.


In this paper, we report on the development and results of our discourse parser for the CoNLL 2015 shared task. Our parser, named \textit{CLaC Discourse Parser}, was built from scratch and took about 3 person-month to code. The focus of the CLaC Discourse Parser is the treatment of explicit discourse relations (i.e. problem 1 to 3 above).

\section{Architecture of the CLaC Discourse Parser}
\label{sec:architecture}

We developed our parser based on the UIMA framework \cite{ferrucci04} and we used ClearTK \cite{bethard14} to add machine learning functionality to the UIMA framework. The parser was written in Java and its source code is distributed under the BSD license\footnote{All the source codes can be downloaded from https://github.com/mjlaali/CLaCDiscourseParser.git}.

Figure~\ref{fig:architecture} shows the architecture of the CLaC Discourse Parser. Motivated by \newcite{lin14}, the architecture of the CLaC Discourse Parser is a pipeline that consists in five components: \textit{CoNLL Syntax Reader}, \textit{Discourse Connective Annotator}, \textit{Argument Labeler}, \textit{Discourse Sense Annotator} and \textit{CoNLL JSON Exporter}. Due to lack of time, we did not implement a \textit{Non-Explicit Classification} in our pipeline and only focused on explicit discourse relations. 

The \textit{CoNLL Syntax Reader} and the \textit{CoNLL JSON Exporter} were added to the CLaC Discourse Parser in order for the input and the output of the parser to be compatible with the CoNLL 2015 Shared Task specifications. The \textit{CoNLL Syntax Reader} parses syntactic information (i.e. POS tags, constituent parse trees and dependency parses) provided by CoNLL organizers and adds this syntactic information to the documents in the UIMA framework. To create a stand-alone parser, the \textit{CoNLL Syntax Reader} can be easily replaced with the \texttt{cleartk-berkeleyparser} component in the CLaC discourse Parser pipeline. This component is a wrapper around the Berkeley syntactic parser \cite{petrov07}  and distributed with ClearTK. The Berkeley syntactic parser was actually used in the CoNLL shared task to parse texts and generate the syntactic information.

The \textit{CoNLL JSON Exporter} reads the output discourse relations annotated in the UIMA documents and generates a JSON file in the format required for the CoNLL shared task. We will discuss the other components in details in the next sections.

\begin{figure*}[!htb]
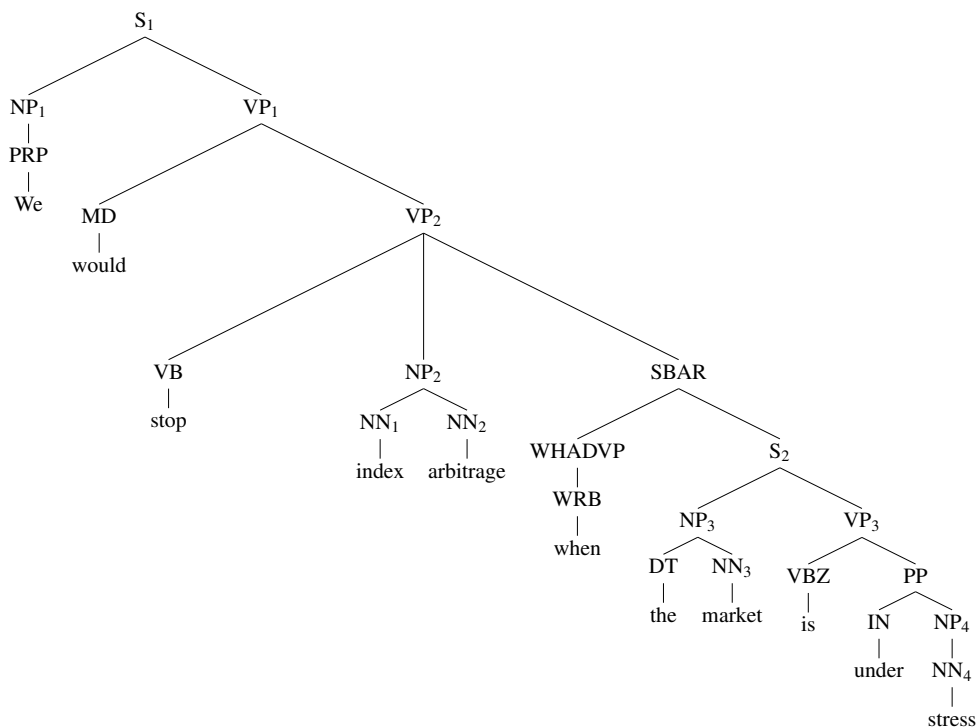

\centering
\resizebox{.8\linewidth}{!}{
\Tree [ .S\textsubscript{1} [ .NP\textsubscript{1} [ .PRP We ] ] [ .VP\textsubscript{1} [ .MD would ] [ .VP\textsubscript{2} [ .VB stop ] [ .NP\textsubscript{2} [ .NN\textsubscript{1} index ] [ .NN\textsubscript{2} arbitrage ] ] [ .SBAR [ .WHADVP [ .WRB when ] ] [ .S\textsubscript{2} [ .NP\textsubscript{3} [ .DT the ] [ .NN\textsubscript{3} market ] ] [ .VP\textsubscript{3} [ .VBZ is ] [ .PP [ .IN under ] [ .NP\textsubscript{4} [ .NN\textsubscript{4} stress ] ] ] ] ] ] ] ] ] }

\caption{The Parse Tree Provided by CoNLL 2015 for Example~\ref{ex:one} }
\label{fig:example}
\end{figure*}

\begin{table*}[!htb]
  \centering
  \begin{tabular}{|p{.15\textwidth}|p{.5\textwidth}|p{.2\textwidth}|}
  \hline
  
    \textbf{Category} & \textbf{Description} & \textbf{Example} \\
    \hline
    \hline
    \multirow{6}{*}{\parbox{.15\textwidth}{Connective Features}} 
    & 1. The discourse connective text in lowercase. & \emph{when} \\ \cline{2-3}
    & 2. The categorization of the case of the connective: \textit{all lowercase}, \textit{all uppercase} and \textit{initial uppercase} & all lowercase \\ \cline{2-3}
    & 3. The highest node in the parse tree that covers the connective words but nothing more & WRB \\ \cline{2-3}
    & 4. The parent of \textit{SelfCat} & WHADVP \\ \cline{2-3}
    & 5. The left sibling of \textit{SelfCat} & null \\ \cline{2-3}
    & 6. The right sibling of \textit{SelfCat} & S \\ \hline
 
    \multirow{3}{*}{\parbox{.1\textwidth}{Syntactic Node Features}} 
    & 7. The path from the node to the \textit{SelfCat} node in the parser tree & $S  	\uparrow SBAR  \downarrow WHADVP$ \\ \cline{2-3}
    & 8. The context of the node in the parse tree. The context of a node is defined by its label the label of its parent, the label of left and right sibling in the parse tree. & S-SBAR-WHADVP-null \\ \cline{2-3}
    & 9. The position of the node relative to the \textit{SelfCat} node: \textit{left} or \textit{right} & left \\ \hline
 
  \end{tabular}
  \caption{Features Used in the CLaC Discourse Parser}
  \label{tab:features}
\end{table*}
\subsection{Discourse Connective Annotator}

To annotate discourse connectives, the \textit{Discourse Connective Annotator} first searches the input texts for terms that match a pre-defined list of discourse connectives. This list of discourse connectives was built solely from the CoNLL training dataset of around 30K explicit discourse relations and contains 100 discourse connectives. Each match of discourse connective is then checked to see if it occurs in discourse usage or not. 

Inspired by \cite{pitler09-a}, we built a binary classifier with six local syntactic and lexicalized features of discourse connectives to classify discourse connectives as discourse usage or non-discourse usage. These features are listed in Table~\ref{tab:features} in the row labeled \textit{Connective Features}.

\subsection{Argument Labeler}

When \textsc{Arg1} and \textsc{Arg2} appear in the same sentence, we can exploit the syntactic tree to label boundaries of the discourse arguments. Motivated by \cite{lin14}, we first classify each constituent in the parse tree into to three categories: part of \textsc{Arg1}, part of \textsc{Arg2} or \textsc{Non} (i.e. is not part of any discourse argument). Then, all constituents which were tagged as part of \textsc{Arg1} or as part of \textsc{Arg2} are merged to obtain the actual boundaries of \textsc{Arg1} and \textsc{Arg2}.

Previous studies have shown that learning an argument labeler classifier when all syntactic constituents are considered suffers from many instances being labeled as \textsc{Non} \cite{kong14}. In order to avoid this, we used the approach proposed by \newcite{kong14} to prune constituents with a \textsc{Non} label. This approach uses only the nodes in the path from the discourse connective (or \textit{SelfCat} see Table~\ref{tab:features}) to the root of the sentence (\textit{Connective-Root path nodes}) to limit the number of the candidate constituents. More formally, only constituents that are directly connected to one of the \textit{Connective-Root path nodes} are considered for the classification.

For example, consider the parse tree of Example~\ref{ex:one} shown in Figure~\ref{fig:example}. The path from the discourse connective \textit{``when''} to the root of the sentence contains these nodes: \{WRB, WHADVP, SBAR, VP\textsubscript{2}, VP\textsubscript{1}, S\textsubscript{1}\}. Therefore, we only consider \{S\textsubscript{2}, NP\textsubscript{2}, VB, MD, NP\textsubscript{1}\} for obtaining discourse arguments.

If the classifier does not classify any constituent as a part of \textsc{Arg1}, we assume that the \textsc{Arg1} is not in the same sentence as \textsc{Arg2}. In such a scenario, we consider the whole text of the previous sentence as \textsc{Arg1}.

In the current implementation, we made the assumption that discourse connectives cannot be multiword expressions. Therefore, the Argument Labeler cannot identify the arguments of parallel discourse connectives (e.g. \textit{either..or}, \textit{on one hand..on the other hand}, etc.)

We used a sub-set of 9 features proposed by \newcite{kong14} for the Argument Labeler classifier. The complete list of features is listed in Table~\ref{tab:features}.

\subsection{Discourse Sense Annotator} 
Although some discourse connectives can signal different discourse relations, the na\"\i ve approach that labels each discourse connective with its most frequent relation performs rather well. According to \newcite{pitler09-a}, such an approach can achieve an accuracy of 85.86\%. Due to lack of time, we implemented this na\"\i ve approach for the Discourse Sense Annotator, using the 100 connectives mined from the dataset (see Section 2.1) and their most frequent relation as mined from the CoNLL training dataset.  

\section{Experiments and Results}
\label{sec:experiments}

\begin{table*}[!htb]
\centering
\begin{tabular}{|l|r|r|r|r|}
\hline
                      & \parbox{.12\textwidth}{\textbf{Discourse \\ Connective \\ Classifier}} & \parbox{.11\textwidth}{\textbf{Argument \\ Labeler}} & \parbox{.2\textwidth}{\textbf{Discourse Parsing \\ (explicit only)}} & \parbox{.2\textwidth}{\textbf{Discourse Parsing \\ (explicit and  \\ implicit)}} \\ \hline
Best Result           & 91.86\%      & 41.35\%          & 30.58\%         & 24.00\%  \\
\textbf{CLaC Parser}           & \textbf{90.19\%}          & \textbf{36.60\%}          & \textbf{27.32\%}        & \textbf{17.38\%} \\
Average               & 74.20\%          & 23.89\%          & 18.28\%        & 13.25\% \\ \hline
Standard deviation    & 23.24\%          & 13.01\%          & 9.93\%         & 6.41\% 
\\ \hline
\end{tabular}
\caption{Summary of the Results of the CLaC Discourse Parser in the CoNLL 2015 Shared Task.}

\label{tab:overall}
\end{table*}

As explained in Section~\ref{sec:architecture}, the CLaC Discourse Parser contains two main classifiers, one for the \textit{Discourse Connective Annotator} and one for the \textit{Argument Labeler}. We used the off-the-shelf implementation of the C4.5 decision tree classifier \cite{quinlan93} available in WEKA \cite{hall09} for the two classifiers and trained them using the CoNLL training dataset.

Although the CLaC discourse parser only considers explicit discourse relations (which only accounts for about half of the relations), the parser ranked 6\textsuperscript{th} among the 17 submitted discourse parsers. The overall F\textsubscript{1} score of the parser and the individual performance of the \textit{Discourse Connective Classifier} and the \textit{Argument Labeler} in the blind CoNLL test data are shown in Table~\ref{tab:overall}. As Table~\ref{tab:overall} shows, the performance of the parser is consistently above the average. In addition, the performance of the \textit{Discourse Connective Classifier} is very close to the best result. 

Note that all numbers presented in Table~\ref{tab:overall} were obtained when errors propagate through the pipeline. That is to say, if a discourse connective is not correctly identified by the \textit{Discourse Connective Classifier} for example, the arguments of this discourse connective will not  be identified. Thus, the recall of the \textit{Argument Labeler} will be affected.

The CoNLL 2015 results of the submitted parsers show that the identification of \textsc{Arg1} is more difficult than \textsc{Arg2}. In line with this, the CLaC Discourse Parser performed better on the identification of \textsc{Arg2} (with the F\textsubscript{1} score of 69.18\%) than \textsc{Arg1} (with the F\textsubscript{1} score of 45.18\%). Table~\ref{tab:arg} provides a summary of the results for the identification of Arg1 and Arg2. An important source of errors in the identification of \textsc{Arg1} is that \textit{attribute spans} are contained within \textsc{Arg1}. For example in~\ref{ex:arg1}, the CLaC Discourse Parser incorrectly includes the text ``\textit{But the RTC also requires ``working'' capital}'' within \textsc{Arg1}. 

\begin{table}[ht!]
\centering
\begin{tabular}{|l|l|l|}
\hline
    & \textbf{Arg1}    & \textbf{Arg2}    \\ \hline
Best Result    & 49.68\% & 74.29\% \\
\textbf{CLaC Parser}    & \textbf{45.18\%} & \textbf{69.18\%} \\
Average & 30.77\% & 50.91\% \\
Std. deviation     & 15.31\% & 20.58\% \\ \hline
\end{tabular}
\caption{Results of the Identification of \textsc{Arg1} and \textsc{Arg2}.}
\label{tab:arg}
\end{table}

\ex. \label{ex:arg1} But the RTC also requires ``working'' capital \textit{to maintain the bad assets of thrifts that are sold} \underline{until} \textbf{the assets can be sold separately}.\footnote{\label{note1}The example is taken from the CoNLL 2015 development dataset.}

With regards to the identification of \textsc{Arg2}, we observed that subordinate and coordinate clauses are an important source of errors. For example in~\ref{ex:arg2}, the subordinate clause ``\textit{before we can move forward}'' is erroneously included in the \textsc{Arg2} span when the CLaC Discourse Parser parses the text. The cause of such errors are usually rooted in an incorrect syntax parse tree that was fed to the parser. For instance in~\ref{ex:arg2}, the text ``\textit{we have collected on those assets before we can move forward}'' was incorrectly parsed as a single clause covered by an S node with the subordinate ``\emph{before we can move forward}'' as a child of this S node. However, in the correct parse tree the subordinate clause should be a sibling of the S node.


\ex. \label{ex:arg2} \textit{We would have to wait} \underline{until} \textbf{we have collected on those assets} before we can move forward.\textsuperscript{\ref{note1}}

\section{Conclusion}
\label{sec:conclusion}

In this paper, we described the CLaC Discourse Parser which was developed from scratch for the CoNLL 2015 shared task. This 3 person-month effort focused on the task of the \textit{Discourse Connective Classification} and \textit{Argument Labeler}. We used a na\"\i ve approach for sense labelling and consider only explicit relations. Yet, the parser achieves an overall F\textsubscript{1} measure of 17.38\%, ranking in 6\textsuperscript{th} place out of the 17 parsers submitted to the CoNLL 2015 shared task.

\section{Acknowledgement}

The authors would like to thank the CoNLL 2015 organizers and the anonymous reviewers. This work was financially supported by NSERC. 

\newpage
\clearpage 

\bibliographystyle{acl}

\end{document}